# *Exploring Integral Image Word Length Reduction Techniques for SURF Detector*

Shoaib Ehsan and Klaus D. McDonald-Maier
School of Computer Science & Electronic Engineering,
University of Essex,
Colchester, UK
sehsan@essex.ac.uk, kdm@essex.ac.uk

***Abstract*—Speeded Up Robust Features (SURF) is a state of the art computer vision algorithm that relies on integral image representation for performing fast detection and description of image features that are scale and rotation invariant. Integral image representation, however, has major draw back of large binary word length that leads to substantial increase in memory size. When designing a dedicated hardware to achieve real-time performance for the SURF algorithm, it is imperative to consider the adverse effects of integral image on memory size, bus width and computational resources. With the objective of minimizing hardware resources, this paper presents a novel implementation concept of a reduced word length integral image based SURF detector. It evaluates two existing word length reduction techniques for the particular case of SURF detector and extends one of these to achieve more reduction in word length. This paper also introduces a novel method to achieve integral image word length reduction for SURF detector.**

## I. INTRODUCTION

Feature extraction is a low level computer vision task consisting mainly of feature detection and description phases. Some generic image features that are usually of interest for extraction are blobs, corners, junctions and contours. Feature extraction serves as primary stage for numerous computer vision applications such as image matching, object recognition and target tracking. Computer vision algorithms are generally computation and data intensive in nature, and feature extraction is no exception. Fast execution speed is, therefore, considered a desirable attribute for any feature extraction algorithm.

A popular algorithm focused on performance enhancement in terms of speed for scale and rotation invariant feature extraction is Speeded Up Robust Features (SURF). According to [1], SURF outperforms its competitors in terms of execution speed with comparable results. For example, it is claimed by [1] that SURF detector is more than three times faster than Difference of Gaussian (DoG) detector employed by SIFT - another state-of-the-art feature extraction algorithm.

The superiority of SURF algorithm in terms of execution speed is based on exploitation of integral image representation for calculation of box type filters during detection and description phases of algorithm [1]. Employment of integral image representation not only eliminates computationally expensive multiplications for box type filter calculation but also reduces computational complexity more by trimming it down to three simple addition operations [2]. Another major advantage that integral image representation brings with it is the ability to calculate box filters of all sizes at constant speed. This is particularly useful for a multi-scale computer vision algorithm like SURF that requires calculation of variable size box filters to implement image pyramid. Calculation time for bigger size filters is thus reduced significantly for SURF leading to speed advantage over other competing algorithms.

The concept of integral image was originally proposed as "Summed-Area Table" by [3] in the field of Computer Graphics. The Viola-Jones face detector was the first one to introduce integral image representation in computer vision domain [2]. Although speed gain and reduced computational complexity are major advantages of utilizing integral image for box type filter calculation, there are some draw backs attached to it. Image pixels are usually represented as 8-bit values. However, a much larger binary word length is required to represent integral image values due to summation. This is a major concern as it directly affects memory size required for storing integral image. As binary word length of integral image is dependent on image size, storage requirements are further increased if bigger size images are involved. A large binary word length for an integral image also implies wider buses and data path elements. Figure 1 depicts binary word length requirement of integral image for some common image sizes. It is assumed here that input image pixels have 8-bit values. A comparison of input image and integral image in terms of memory requirements is shown in Figure 2. The percentage increase in storage requirements from input image to integral image is given in Figure 3 for some common image sizes. It can be clearly seen from these figures that increasing input image size has adverse effect on binary word length and memory requirements for integral image. This is not desirable and needs to be addressed accordingly.

This paper proposes a novel implementation concept for a SURF detector that is based on a reduced word length integral image to achieve minimization of hardware resources. Two existing techniques, originally presented for achieving integral image word length reduction in Viola-Jones face detector [4], are evaluated for use with the SURF detector in this paper. One of these techniques is extended by this paper to achieve further word length reduction for SURF detector. This is followed by the introduction of a novel word length reduction method for integral images in a SURF detector.

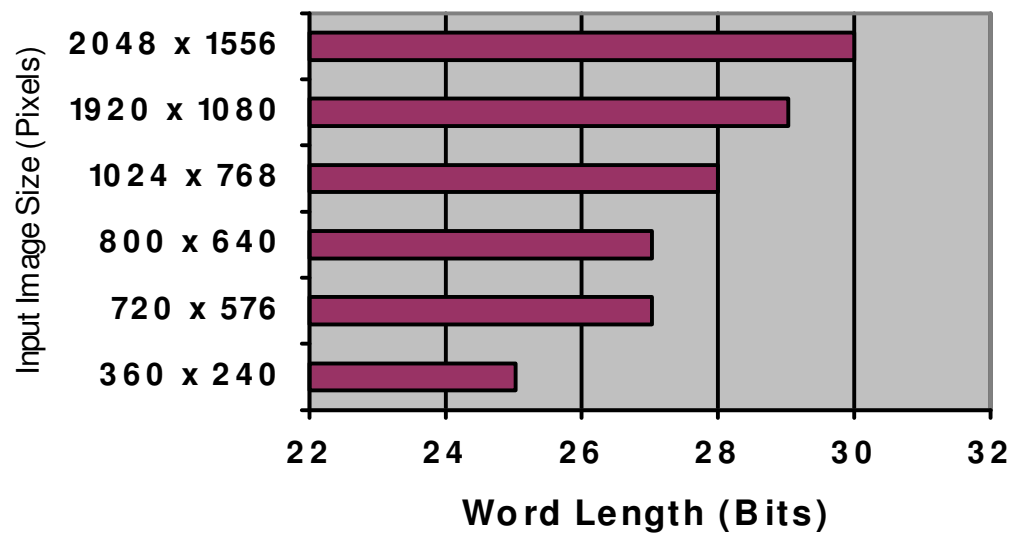

Figure 1. Binary word length requirement for integral image for some common image sizes

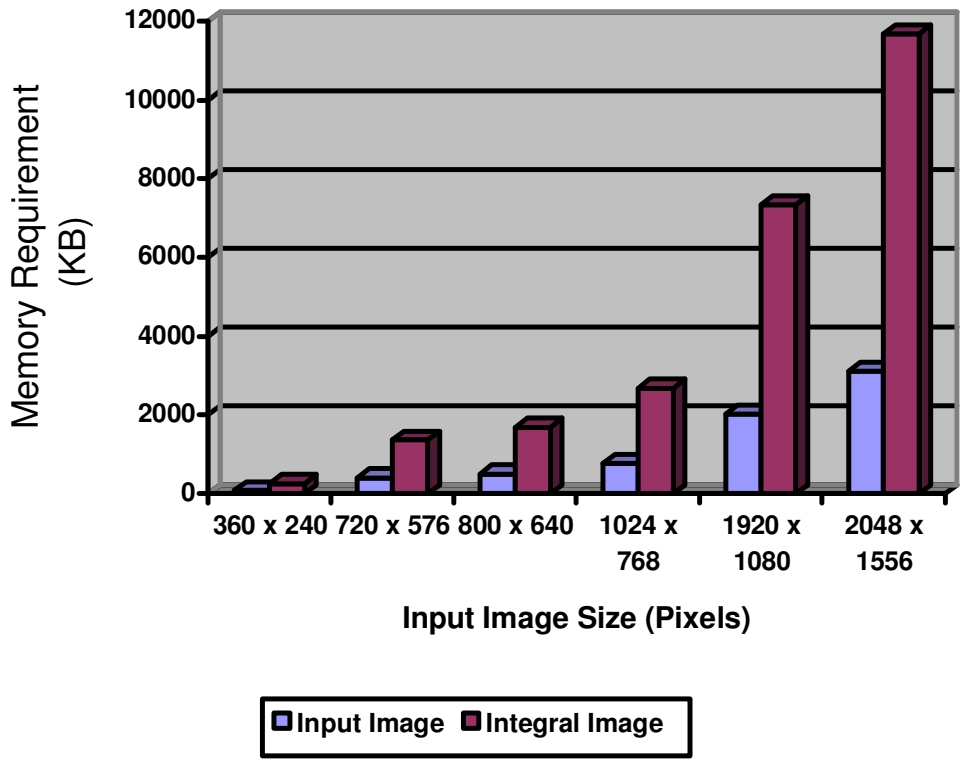

Figure 2. Comparison of Memory requirements for input image and integral image for some common image sizes

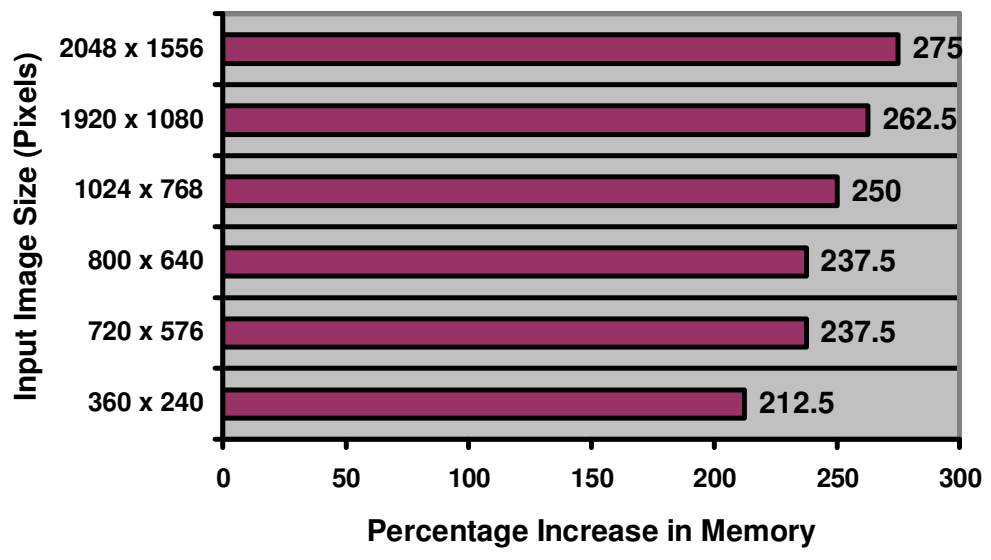

Figure 3. Percentage increase in memory requirement from input image to integral image for some common image sizes

## II. INTEGRAL IMAGE: A RAPID BOX FILTER CALCULATOR

Integral image is an intermediate representation for an image that allows fast calculation of rectangle features [2]. The size of integral image is the same as the input image in terms of pixels. Value of integral image at any location $(x,y)$ is sum of all the pixels to the left and above it in input image including itself. This can be stated mathematically by the following equation:

$$ii(x,y) = \sum_{x' \le x, y' \le y} i(x',y') \quad (1)$$

where $ii(x,y)$ and $i(x,y)$ are values of integral image and input image respectively at location $(x,y)$.

The utilization of the integral image instead of an original input image in an image processing/computer vision application implies computational overhead due to the calculation of integral image. However, this overhead is justified by significant reduction in computational complexity of box type filters which is trimmed down to only three addition operations. Moreover, box type filters of any size can be computed at constant speed using integral image which is advantageous for multi-scale image processing algorithms. A box type filter, as shown in Figure 4 by gray area, can be calculated using integral image as given by the following equation:

$$\begin{aligned} BoxFilterValue &= ii(x_e, y_e) - ii(x_e, y_o - 1) \\ &- ii(x_o - 1, y_e) + ii(x_o - 1, y_o - 1) \end{aligned} \quad (2)$$

Integral image values that are used in Equation 2 for calculation of box filter are shown in black color in Figure 4.

Employment of integral image representation has a negative impact on memory requirements. The worst case integral image value that determines the binary word length required to represent integral image is dependent upon the width, height and number of bits per pixel of input image. This can be stated mathematically as given in [4]:

$$ii_{max} = (2^{L_i} - 1)\, W H \quad (3)$$

where *'i'* is input image, *'ii'* is integral image, *'W'* is width of input image, *'H'* is height of input image and $'L_i'$ is number of bits per pixel of input image. According to [4], the number of bits $'L_{ii}'$ required for representing worst case integral image value needs to satisfy

$$(2^{L_{ii}} - 1) \ge (2^{L_i} - 1)\, W H \quad (4)$$

Finally, total memory in bytes required to store integral image can be calculated as follows:

$$Memory = (WH)L_{ii} / 8 \quad (5)$$

## III. REVIEW OF SURF DETECTOR

This section provides a brief review of SURF detector. An in depth discussion of SURF detector can be found in [1]. The SURF detector makes use of integral image representation to enhance performance in terms of execution speed and reduce computational complexity. The first stage is the calculation of the integral image representation for an input image according to Equation 1. The next step involves calculation of blob response maps at different scales to implement image pyramid for scale-space analysis. The Scale-space is divided into number of octaves that are formed by grouping blob response maps for adjacent scales.

Blob response at any scale 'σ' for any point x = (x,y) in an input image 'I' is calculated using the following equation:

$$\det(H_{approx}) = D_{xx}D_{yy} - (0.9D_{xy})^2 \quad (6)$$

where $D_{xx}$, $D_{yy}$ and $D_{xy}$ are convolution of input image with approximate second order Gaussian partial derivative in $x$, y and xy direction respectively centered at point x = $(x,y)$.

These blob responses are also normalized with respect to the mask size. Second order Gaussian partial derivatives used for calculation of $D_{xx}$, $D_{yy}$ and $D_{xy}$ are approximated as box filters in SURF algorithm as shown in Figure 5. Since the SURF detector implements image pyramid for scale-space analysis by increasing the filter size rather than sub-sampling original image, employing an integral image representation allows the fast calculation of $D_{xx}$, $D_{yy}$ and $D_{xy}$ using Equation 2 for any box filter size at constant speed. Once the blob response maps at different scales are calculated, 3-D Non-Maximum Suppression is carried out to determine the local maxima. In order to achieve sub-pixel, sub-scale accuracy, 3-D quadratic interpolation is done that provides the required interpolated interest points.

## IV. Reduced Word Length Integral Image Based SURF Detector

With the objective of minimizing hardware resources, this paper presents a novel idea of SURF detector implementation based on reduced word length integral images. Previously, a reduced word length integral image concept was successfully applied to Viola-Jones face detector in [4]. This paper evaluates word length reduction techniques presented in [4] for a SURF detector implementation, extends one of these techniques to achieve more word length reduction and introduces a novel approximate technique for word length reduction of integral image in SURF detector.

Before discussing techniques employed for word length reduction, it is worth mentioning here that an implementation of SURF algorithm presented in [1] was done in MATLAB and was successfully tested using data sets provided by [5]. As an example, for the first image of Graffiti scene, our code detects 1421 interest points with a threshold of 50000 where as binary provided by [6] detects 1411 points as mentioned in [1]. This MATLAB code was then modified to test different word length reduction techniques for SURF detector. A discussion of word length reduction techniques is given below.

### A. *The Exact Method*

This technique was presented by [4] for platforms with complement-coded arithmetic and tested for the Viola-Jones face detector. According to this method, if a chain of linear operations is performed on integers and there are some intermediate overflowing results then it is possible to get the correct final result if this result can be represented by the used data word length. The primary advantages of this

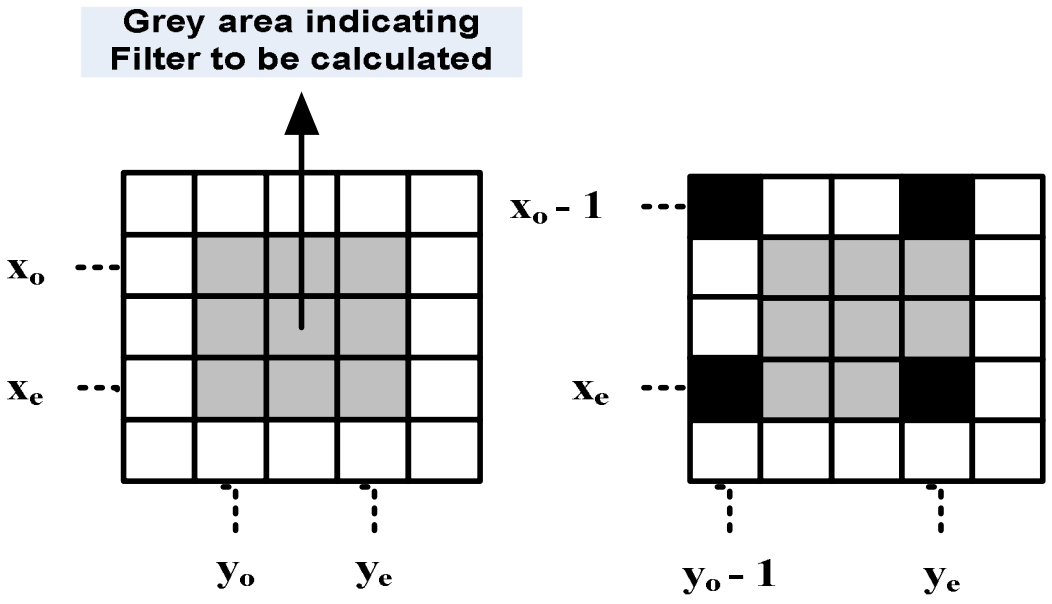

Figure 4. Box type filter calculation using integral image

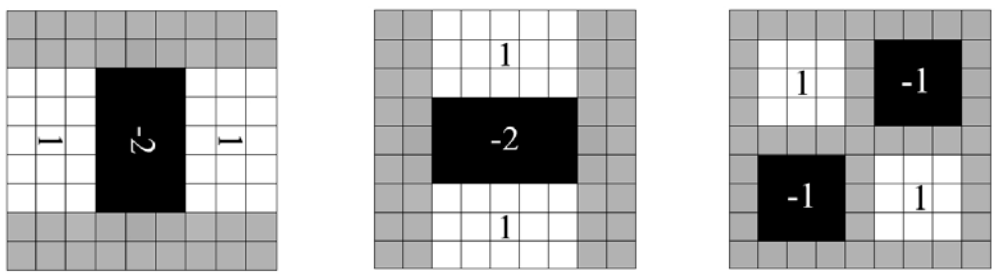

Figure 5. Box filter approximations for second order Gaussian partial derivatives in *x*-direction, *y*-direction and *xy*-direction (left to right). The gray regions are equal to zero [1]

technique are that there is no loss of accuracy and no extra operations are required.

The Exact method is applicable only in situations where the maximum size of box filter to be calculated is a priori known and is smaller than input image size. The SURF detector satisfies this criterion as it requires the box filter size to be less than or equal to the size of input image minus three samples in order to apply at least one non-maximum suppression iteration [7]. Since this analysis intends to determine worst case memory and word length requirements for integral image with the Exact method for SURF detector, it is important to consider a high resolution image. Thus, it is assumed here that the input image size is 800 x 640.

An important fact that needs to be considered while determining the maximum size of box filter for SURF detector is that the number of detected interest points per octave decay very quickly [1]. Therefore, it is sufficient to have only four octaves for scale-space analysis of SURF detector with the biggest filter size being 195 x 195 for all images equal to or greater than this filter size. This implies that the results presented here are not only limited to an image size of 800 x 640 but also apply to any image size equal to or bigger than 195 x 195.

The 195 x 195 filter can be broken down into three box type filters of 65 x 129 each for calculation of $D_{xx}$ and 129 x 65 each for calculation of $D_{yy}$. Hence, the maximum box filter size that needs to be calculated for SURF detector is 129 x 65(or 65 x 129). According to [4], if maximum height and width of box filter to be calculated are known, then the word length for the integral image needs to satisfy:

$$(2^{L_{ii}} - 1) \geq (2^{L_i} - 1)\, W_{max}\, H_{max} \quad (7)$$

where $W_{max}$ = Maximum width of box filter
$H_{max}$ = Maximum height of box filter

For the SURF detector, when assuming 8-bit input image pixels, the worst case word length for the integral image using the Exact method is 22-bits for all input image sizes equal to or greater than 195 x 195 pixels. For example, the word length requirement for integral image is significantly reduced from 27-bits to 22-bits for an image size of 800 x 640. The memory requirement of integral image is also significantly reduced from 1687.5 Kbytes to 1375 Kbytes for an image size of 800 x 640. The percentage reduction in storage requirements for integral image with the Exact method as compared to the full word length case for some common image sizes in the case of SURF detector is shown in Figure 6. The SURF detector was tested successfully with 22-bit integral image for all image datasets with out any loss of accuracy.

### *B. The Modified Exact Method*

This paper extends the Exact method by noting that the assumption of having all pixels in input image set to their maximum value (255 in case of 8-bit pixels) for evaluating the worst case integral image value does not seem very practical for SURF detector which is used for blob detection, i.e. to detect dark areas/regions in image surrounded by light ones or vice versa. Assuming that all the pixels are set to their maximum value in input image implies that there are no blobs to be detected by SURF in the input image. The technique presented by [4] is, therefore, further extended in this paper by assuming that there is variation in pixel values of input image and modifying the Inequality 7 as:

$$(2^{L\,ii} - 1) \geq (2^{L\,i} - 1)\,(W_{max}\,H_{max})\,(0.96) + (2^{L\,i-1} - 1)\,(W_{max}\,H_{max})\,(0.04) \quad (8)$$

It is assumed here that 96 % of all pixels in a box filter to be evaluated have maximum values where as only 4 % pixels have half the maximum value. This is a suitable approximation here as most images generally have more variation in pixel values than given by Inequality 8. This method allows the word length of integral image to be reduced to 21-bits and the approximation used in this method was justified by successful testing of SURF detector without any loss of accuracy for all the image datasets. As an example, for the first scene of Graffiti image, same number of interest points (1421) was detected with 21-bit word length of integral image by SURF detector. Percentage reduction in storage requirements for integral image with modified exact method as compared to full word length case for some common image sizes in the case of SURF detector is shown in Figure 7. To further ensure that the final result is correct and does not overflow, an overflow flag can also be used when designing hardware.

Loss of accuracy was observed when word length was reduced below 21-bits using modified exact method for image data sets. Table I provides detailed data for Graffiti data set and indicates inaccuracy of SURF detector with 20-bit integral image.

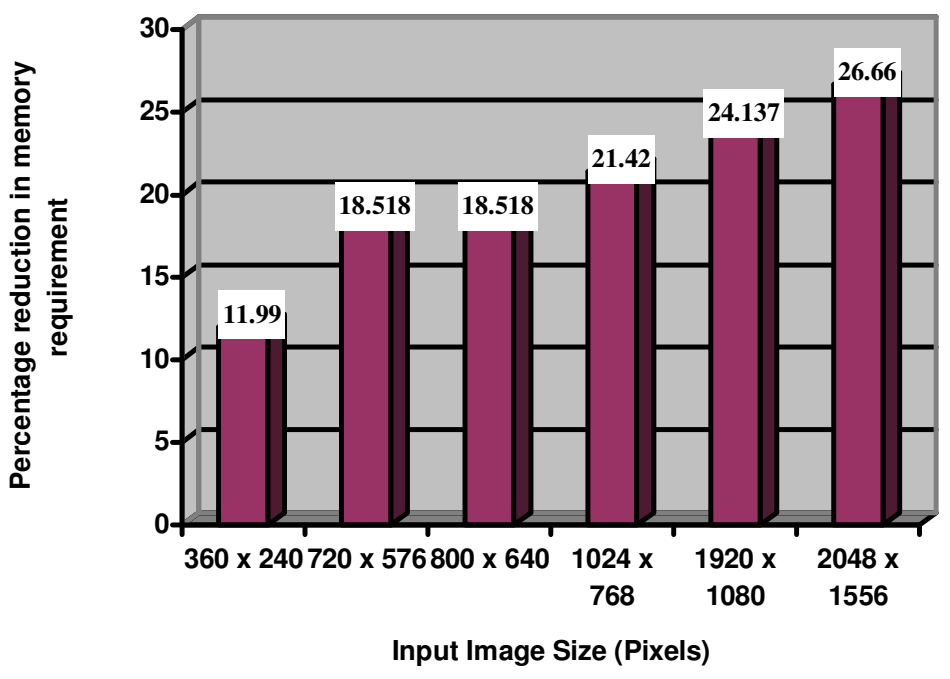

Figure 6. Percentage reduction in integral image storage requirements with the Exact method for SURF Detector

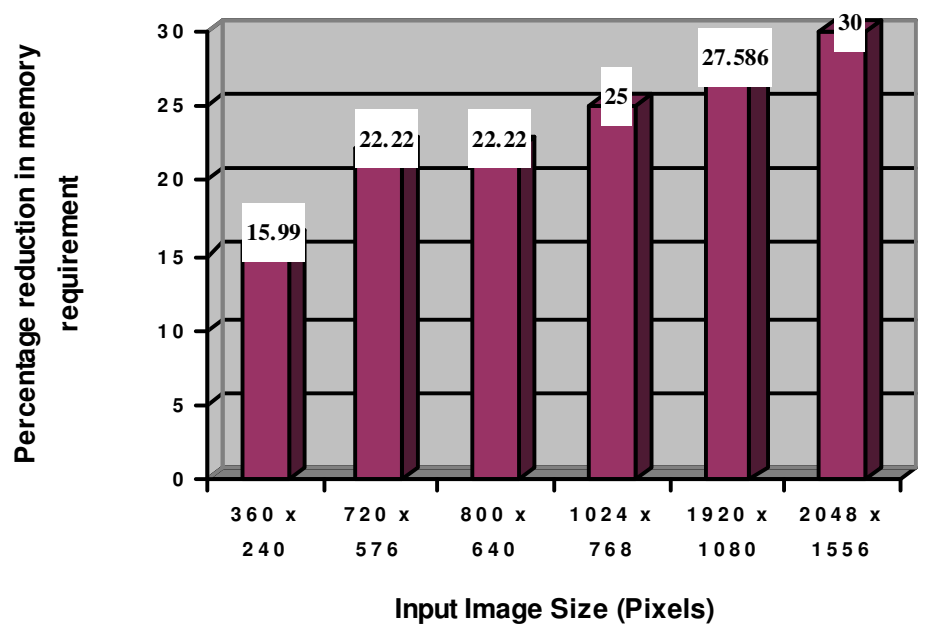

Figure 7. Percentage reduction in integral image storage requirements with Modified Exact Method for SURF detector

TABLE I. LOSS OF ACCURACY OF SURF DETECTOR BASED ON 20-BIT INTEGRAL IMAGE FOR GRAFFITI IMAGE DATA SET

| Image Number | Interest Points detected by SURF Detector | |
|---|---|---|
| | Full word length integral image | 20-bit Integral Image |
| 1 | 1421 | 1414 |
| 2 | 1464 | 1496 |
| 3 | 1439 | 1429 |
| 4 | 1193 | 1189 |
| 5 | 1175 | 1210 |
| 6 | 870 | 911 |

### *C. The Approximate Method*

This method was successfully applied to the Viola-Jones face detector and an in-depth detail can be found in [4]. In this method, the input image is rounded by shifting out pixel bits. Thus, the number of bits per pixel '$L_i$' of input image is reduced and results in a smaller word length for integral image as compared to the Exact method. However, this rounding of input image introduces an error which is reduced by distributing the rounding error across several neighboring pixels.

This method was implemented for the SURF detector but experiments showed highly inaccurate results due to the rounding of input image pixels. For example, one-bit shifting out of pixel values reduced integral image word length to 21-bits but it affected the accuracy of SURF detector severely reducing the number of detected interest points to 212 for the first image of Graffiti scene and was unable to detect any interest points for bit shifts greater than 1. Moreover, this method also involves extra addition operations for the calculation of the reduced word length integral image which is not desirable.

## D. *The Even Image Method*

This paper presents a novel approximate technique for word length reduction of integral image in a SURF detector. This technique adds 1-bit noise to those input image pixels that have odd values to make the whole image even. The advantage of making the image even-valued is that if a specific number of bits are shifted out from an even pixel and the value of pixel remains non-zero after the shift then its original value can be recovered with the same amount of left shift at a later stage. For example, if a pixel has value '4' with binary word '100', then shifting out of 2 bits will make its value '1' with binary word '001'. The original value of pixel can then easily be recovered by having a left shift of 2-bits at some later stage. Pixels with odd values do not possess this property and hence are converted to even values in this particular method. This 'Even' input image is then rounded by shifting out 'p' bits from each pixel. The reduced word length integral image is then calculated as in the Exact method. When this reduced word length integral image is used for box filter calculation according to Equation 2, the final result is shifted to the left by 'p' bits to get the correct box filter value.

This method was tested using all image datasets and the results for Graffiti data set are shown in Table II for 'p' bit shift. The integral image word length '*Lii*' has been calculated in Table II by using '*Li – p'* instead of '*Li*' in modified exact method equation. It can be seen that this method performs significantly better in terms of accuracy of the SURF detector as compared to the Approximate method and allows the integral image word length to be reduced to 19-bits with fairly accurate results. The Percentage reduction in storage requirements for integral image with the Even image method for a word length of 19-bits as compared to full word length case for some common image sizes in the case of SURF detector is shown in Figure 8. The Even image method suffers from inability to recover pixel values that become zero due to the initial truncation. This is the reason that the inaccuracy of the detector increases with increasing amount of shift. Extra addition operations to make input image even and additional shift operations to get the correct value for box filter are some draw backs of this approach.

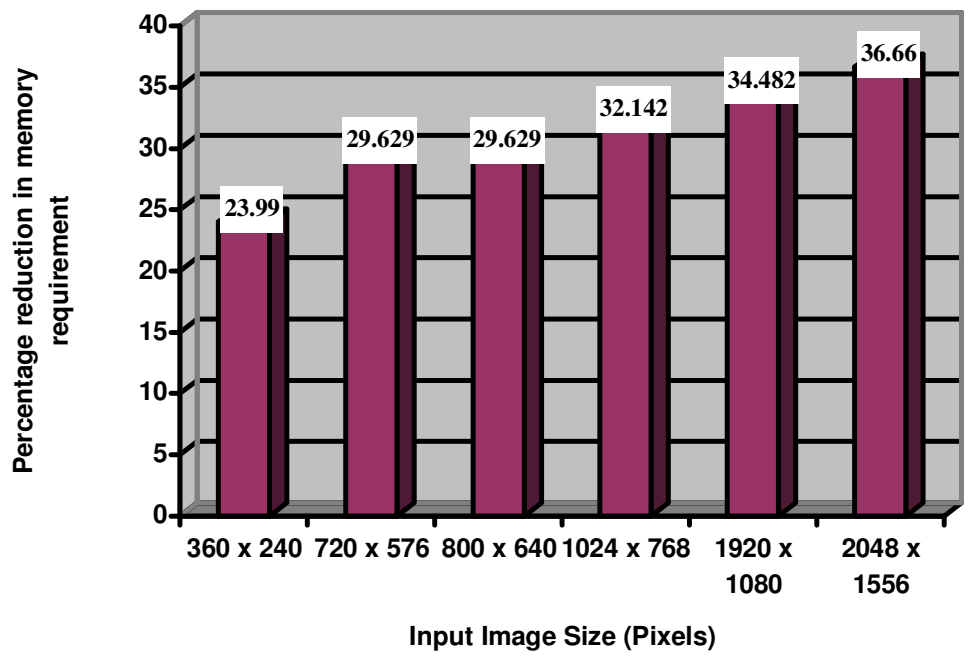

Figure 8. Percentage reduction in integral image storage requirements with Even Image Method for SURF detector

TABLE II. RESULTS OF SURF DETECTOR WITH THE EVEN IMAGE METHOD FOR GRAFFITI DATA SET

| p | $L_{ii}$ | Interest Points detected by SURF Detector for Graffiti data set | | | | | |
|---|---|---|---|---|---|---|---|
| | | Img 1 | Img 2 | Img 3 | Img 4 | Img 5 | Img 6 |
| 1 | 20 | 1413 | 1482 | 1433 | 1197 | 1176 | 877 |
| 2 | 19 | 1421 | 1478 | 1419 | 1192 | 1170 | 868 |
| 3 | 18 | 479 | 505 | 479 | 381 | 428 | 299 |
| 4 | 17 | 20 | 29 | 16 | 27 | 33 | 26 |

# V. CONCLUSIONS

This paper has presented a novel implementation concept for the SURF detector which is based on a reduced word length integral image. Two existing word length reduction techniques for integral image have been evaluated for achieving significant reduction in memory requirements and width of data path elements for SURF detector. This paper has also extended the Exact method presented in [4] for SURF detector and presented a novel method to achieve greater reduction in memory requirements of an integral image.

Future work includes investigation of parallel methods for computation of integral image in hardware. Based on the work presented in this paper, a novel reduced word length, parallel architecture for integral image calculation will be designed for SURF and prototyping will be done on FPGA.

# REFERENCES

[1] Bay, H., Tuytelaars, T. and Gool, Luc V., "Speeded Up Robust Features (SURF)," in Computer Vision and Image Understanding, Vol. 110, No. 3, pp. 346-359, June 2008.

[2] Viola, P. and Jones, M., "Rapid Object Detection using a Boosted Cascade of Simple Features," in Proceedings of CVPR 2001, pp. 511-518.

[3] Crow, F., "Summed-Area Tables for Texture Mapping," in Proceedings of SIGGRAPH, Vol. 18, No. 3, pp. 207-212, 1984.

[4] Belt, H.J.W., "Word Length Reduction for the Integral Image" in Proceedings of IEEE International Conference on Image Processing 2008, pp. 805-808.

[5] http://www.robots.ox.ac.uk/~vgg/research/affine/, accessed Jan 09, 2009.

[6] http://www.vision.ee.ethz.ch/~surf/download_ac.html, accessed Jan 05, 2009.

[7] SURF Patent document, published Nov 15, 2007.